\documentclass[conference]{IEEEtran}
\IEEEoverridecommandlockouts
\usepackage{booktabs}
\usepackage{makecell}
\usepackage{cite}
\usepackage{amsmath,amssymb,amsfonts}
\usepackage{graphicx}
\usepackage{textcomp}
\usepackage{xcolor}
\usepackage{multirow}
\usepackage{array}
\usepackage{float}
\usepackage{comment}
\def\BibTeX{{\rm B\kern-.05em{\sc i\kern-.025em b}\kern-.08em
    T\kern-.1667em\lower.7ex\hbox{E}\kern-.125emX}}

\begin{document}

\title{A Self-Consistency-Based Reranking for Narrative Question Answering}

\author{
\IEEEauthorblockN{Molham Mohamed}
\IEEEauthorblockA{\textit{Faculty of Computer Science} \\
\textit{MSA University} \\
Giza, Egypt \\
molham.mohamed@msa.edu.eg}
\and
\IEEEauthorblockN{Ali Hamdi}
\IEEEauthorblockA{\textit{Faculty of Computer Science} \\
\textit{MSA University} \\
Giza, Egypt \\
ahamdi@msa.edu.eg}
}

\IEEEpubid{\makebox[\columnwidth]{979-8-3315-8488-7/26/\$31.00 ©2026 IEEE \hfill}
\hspace{\columnsep}\makebox[\columnwidth]
{}}
\maketitle

\begin{abstract}
Narrative question answering (NQA) is a challenging task in natural language processing that requires models to understand long textual contexts, capture relationships across events, and generate coherent responses. Despite recent advances in pretrained language models, most existing approaches rely on a single decoding output during inference, making them sensitive to generation variability and often resulting in incomplete or inconsistent answers.To address this limitation, we propose a self-ensemble Self-Consistency-Based reranking framework for narrative question answering. The proposed method generates multiple candidate answers for each story-question pair and selects the final answer based on semantic agreement among the generated responses. This allows the model to explore diverse answer formulations while improving robustness through consensus-based selection without requiring modifications to the underlying architecture.The framework combines pretrained and fine-tuned language generation with multi-answer inference and similarity-based reranking. We evaluate the proposed approach on the NarrativeQA dataset using multiple models, including FLAN-T5 (Base and Small) and Pegasus-Large, under both baseline and fine-tuned settings.Experimental results demonstrate that the proposed method consistently improves performance across all models. In particular, FLAN-T5-Base achieves the best overall performance, improving from 82.32\% to 86.66\% (+4.34\%) when combined with self-ensemble inference. Additionally, the largest improvement is observed with Pegasus-Large, which increases from 72.50\% to 87.07\% (+14.57\%), highlighting the effectiveness of the proposed strategy.

\end{abstract}


\section{Introduction}

Narrative question answering (NQA) is a challenging task in natural language processing that requires models to understand long narrative contexts and generate accurate and coherent responses. Unlike extractive question answering, NQA involves reasoning over complex story structures, including events, characters, and temporal dependencies. This makes the task inherently difficult, as models must capture both deep contextual understanding and generative capabilities to produce semantically correct answers.

Recent advances in transformer-based language models have significantly improved performance in generative question answering. However, most existing approaches rely on a single decoding trajectory during inference, making the final output highly sensitive to stochastic decoding and limiting robustness. In practice, different decoding runs for the same input may produce different answers, some of which may be incomplete or inconsistent.

To address this issue, generating multiple candidate answers has emerged as an effective strategy. Instead of relying on a single output, the model produces multiple responses for the same input, allowing it to explore different possible answers. This idea is closely related to self-consistency, where agreement among multiple generated outputs is used as a signal of correctness \cite{wang2022selfconsistency, ahmed2023selfconsistency , chang2024ensemble, xu2025selfensemble}. In general, answers that appear consistently across different generations are more likely to be reliable.

However, most existing approaches rely on exact matching or majority voting when selecting the final answer. This assumption is often too strict for open-ended generation tasks, where semantically similar answers may differ in wording or structure. As a result, correct answers may not be selected simply because they do not match exactly with other candidates.

In this work, we address this limitation by introducing a semantic self-consistency-based reranking strategy. Instead of comparing answers at the surface level, the proposed method evaluates agreement at the semantic level. This allows the framework to capture similarity in meaning rather than exact wording, leading to more flexible and robust answer selection \cite{hamdi2024riro}.

The proposed framework follows a self-ensemble approach, where multiple candidate answers are generated and then compared using semantic similarity. The final answer is selected based on its overall agreement with the other candidates. This idea is supported by prior work showing that consistency across multiple predictions can improve reliability and reduce uncertainty \cite{golestaneh2020consistency, hawkins2021consistency}. In addition, ensemble-based reasoning has been shown to improve both accuracy and confidence by combining multiple outputs \cite{chang2024ensemble, xu2025selfensemble, elgabry2025confidence, hamdi2025ensemble}.

\begin{figure*}[t]
\centering
\includegraphics[width=0.7\linewidth,height=4.5cm]{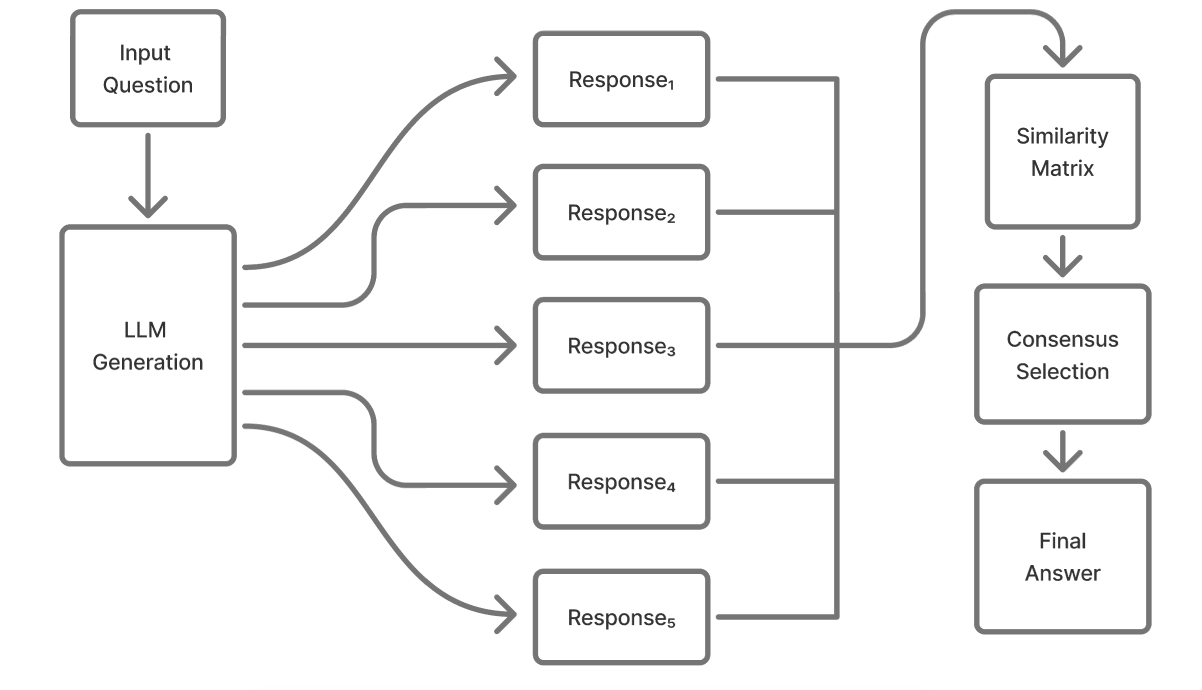}
\caption{Overview of the proposed self-ensemble framework. The language model generates multiple candidate responses, which are compared using semantic similarity. A self-consistency-based selection mechanism then identifies the most consistent response as the final answer.}
\label{fig:diagram}
\end{figure*}

As illustrated in Fig.~\ref{fig:diagram}, the model first generates multiple candidate responses for a given input question. These responses are then compared using semantic similarity, and a consensus-based mechanism selects the most consistent answer as the final output.

Beyond basic consistency, recent work has explored improving inference through more advanced strategies such as adaptive sampling, reasoning-aware selection, and multi-step refinement \cite{wan2025rasc, wan2024dynamic, wang2025srnle}. These approaches further highlight that improving inference-time decision making is essential for reliable generation.

Despite these advances, most existing methods still rely on surface-level agreement or model-based scoring, which may not generalize well to free-form generation tasks. In contrast, the proposed approach focuses on semantic agreement, providing a more flexible and task-appropriate solution.

To further support the effectiveness of semantic similarity, we leverage high-quality sentence embeddings, which have been shown to capture meaningful relationships between text representations \cite{gao2021simcse, akl2025fusion}. This allows the model to compare candidate answers more accurately and select the most consistent one.

Overall, this work proposes a simple yet effective framework that improves robustness in narrative question answering by combining multi-answer generation with semantic consistency-based reranking.

The remainder of this paper is organized as follows. Section II reviews related work in generative question answering and consistency-based inference strategies. Section III presents the proposed methodology. Section IV reports experimental results and discussion. Finally, Section V concludes the paper.

\section{Related Work}

Generative question answering requires models to produce answers based on contextual understanding rather than direct extraction. While pretrained transformer-based models have significantly improved performance, the quality of generated responses remains highly dependent on inference strategies. In particular, relying on a single decoding path often leads to unstable outputs due to stochastic generation and limited exploration of possible answers \cite{wang2023selfconsistency}. This limitation highlights the need for more robust inference mechanisms that can better handle variability in generation.

To address this issue, multi-candidate generation has been introduced as an effective alternative. Instead of producing a single answer, the model generates multiple candidates for the same input, allowing it to explore different reasoning paths and answer formulations. This idea is closely related to self-consistency, where the final answer is selected based on agreement among multiple generated responses \cite{wang2022selfconsistency, chen2023usc}. Such approaches reduce the impact of randomness and improve reliability by leveraging consensus among outputs. Moreover, they demonstrate that correct answers tend to appear consistently across diverse samples.

A deeper understanding of self-consistency has been provided by analysing it as a distributional process. The generated answers can be viewed as samples from an underlying answer distribution, where decoding parameters such as temperature influence both diversity and convergence \cite{li2025distributional}. This introduces a trade-off between exploration and stability. Increasing diversity allows better coverage of possible solutions but requires more effective aggregation to filter inconsistent outputs, while low diversity may reinforce biased predictions. Adaptive methods have therefore been proposed to dynamically adjust sampling behaviour and improve efficiency under limited sampling budgets \cite{zhou2025rpc}.

Beyond standard self-consistency, several approaches focus on improving answer selection using additional evaluation signals. Verification-based methods introduce auxiliary scoring mechanisms to assess candidate outputs and filter incorrect responses \cite{lightman2023verify}. Similarly, reasoning-aware frameworks evaluate intermediate reasoning steps alongside final answers, enabling more informed selection of candidates \cite{wan2025rasc}. Dynamic self-consistency methods further extend this idea by incorporating adaptive stopping and weighted aggregation strategies to reduce computational cost while maintaining performance \cite{wan2024dynamic}. These approaches highlight that effective answer selection often requires combining multiple signals rather than relying solely on agreement.

Another important direction involves ensemble-based reasoning, where multiple reasoning processes or outputs are combined to improve decision reliability. These methods demonstrate that aggregating diverse reasoning signals can lead to more consistent and accurate results compared to relying on a single inference path \cite{chang2024ensemble, xu2025selfensemble, elgabry2025confidence, hamdi2025ensemble}. In addition to improving accuracy, such approaches also enhance confidence estimation by aligning agreement with prediction reliability.

In parallel, training strategies have been explored to improve generation quality. Severity-aware learning introduces weighted loss functions that assign different importance levels to prediction errors, allowing models to focus on more critical mistakes \cite{molham2026severity}. Curriculum learning further improves model performance by presenting training data in a structured order, typically progressing from simpler to more complex examples \cite{molham2026curriculum}. These approaches improve learning efficiency and help models better capture complex patterns. However, they primarily operate during training and do not directly address variability at inference time.

Consistency has also been studied as a general learning principle beyond language modeling. Enforcing agreement across multiple predictions has been shown to improve reliability and reduce uncertainty in different domains, including semantic segmentation and graph-based learning \cite{golestaneh2020consistency, hawkins2021consistency}. These findings suggest that consistency can serve as a general indicator of prediction quality across tasks.

More recently, collaborative reasoning strategies have been explored to improve consistency through structured interaction. Multi-agent approaches allow multiple model instances to exchange and refine their reasoning, producing stronger consensus signals compared to independent sampling \cite{samanta2026maca}. This demonstrates that interaction between reasoning processes can further enhance robustness in complex tasks.

Another line of work focuses on iterative refinement, where models improve their outputs through repeated feedback and correction. Self-refinement methods enable models to identify weaknesses in initial responses and progressively enhance them, leading to more accurate and faithful outputs \cite{wang2025srnle}. This approach complements multi-sample inference by improving the quality of candidate answers before selection.

Self-consistency has also been applied in practical settings such as program repair, where multiple candidate solutions are generated and the most frequent output is selected. These results provide further evidence that correct solutions tend to emerge consistently across different reasoning paths \cite{ahmed2023selfconsistency}. This reinforces the effectiveness of aggregation-based reasoning strategies in real-world applications.

Overall, prior work highlights that generating multiple candidate outputs and selecting among them is a powerful approach for improving reliability in generative models. However, most existing methods rely on exact matching, majority voting, or model-based scoring and reranking mechanisms \cite{aloraini2025lexisem, elewa2025balancing}. These approaches may not generalize well to open-ended generation tasks, where semantically similar answers can differ in surface form. This limitation motivates the need for semantic-level aggregation strategies that can better capture agreement among diverse outputs while preserving flexibility in generation.

\section{Methodology}

This study proposes a Self-Ensemble Self-Consistency-Based Reranking Strategy for narrative question answering. 
The proposed framework improves answer quality by combining task-specific fine-tuning with multi-answer generation and semantic similarity-based reranking.

The overall framework consists of five main components: dataset preparation, preprocessing, model selection, fine-tuning, and consensus-based answer selection. Each component is designed to address a specific limitation in generative question answering, particularly the instability introduced during inference.

The framework focuses on improving both the quality of generated answers and the reliability of the final selection process. While fine-tuning enhances the model's ability to produce relevant outputs, the self-ensemble mechanism ensures that the final answer is selected based on agreement among multiple candidates.

\subsection{Dataset Description}

The experiments are conducted on the NarrativeQA dataset, which contains story summaries, questions, and corresponding reference answers. Each sample is represented as a triplet $(s, q, a)$, where $s$ denotes the story summary, $q$ is the question, and $a$ is the ground-truth answer.

The dataset is designed to evaluate narrative understanding, requiring models to process long contexts and capture relationships between events and entities. Unlike short-context datasets, NarrativeQA introduces additional complexity due to the length and structure of the input text.

The task is formulated as a conditional text generation problem in which the model generates an answer $a$ given the input pair $(s, q)$. The dataset is divided into training and testing subsets, where the training set is used for supervised fine-tuning and the test set is used for evaluation.

This setup allows the model to learn task-specific patterns during training while ensuring that evaluation is performed on unseen data. As a result, the reported performance reflects the generalization ability of the model.

\subsection{Preprocessing}

To ensure consistency, each sample is transformed into a unified input--output format. The story summary and question are concatenated into a single input sequence, while the corresponding answer is used as the target output.

Basic preprocessing steps are applied, including removing empty entries and ensuring valid text formatting. These steps improve training stability and reduce noise in the dataset.

In addition, consistent formatting helps the model better understand the structure of the input, especially when combining multiple text components such as the story and the question. This improves the alignment between input and output during training.

These preprocessing steps also ensure reliable evaluation, particularly when automatic metrics such as BERTScore are used, as they depend on clean and well-structured text.

\subsection{Problem Formulation}

Let
\[
D = \{(s_i, q_i, a_i)\}_{i=1}^{N}
\]
denote the dataset. The objective is to learn a conditional generation function:
\[
f(s, q) \rightarrow \hat{a}
\]
that produces an answer $\hat{a}$ given a story-question pair $(s, q)$.

This formulation treats narrative question answering as a sequence-to-sequence learning problem, where the model maps textual input to a generated output. The goal is not only to produce grammatically correct answers, but also to ensure semantic correctness.

The model must therefore capture both local context (within sentences) and global context (across the entire story), which makes the task more complex than standard text generation problems.

\subsection{Model Selection}

The proposed framework employs FLAN-T5 as the backbone language model. FLAN-T5 is a pretrained encoder--decoder transformer that has demonstrated strong performance in instruction-following and generative tasks.

Its architecture is well suited for conditional generation, where the model maps structured textual input to a target output sequence. The encoder processes the input text, while the decoder generates the answer token by token.

Two variants are considered: FLAN-T5-Base and FLAN-T5-Small. These models allow evaluation across different model sizes to study the effect of capacity on performance.

In addition, Pegasus-Large is used to further validate the generality of the proposed approach across different sequence-to-sequence models. This ensures that the method is not limited to a single architecture.

\subsection{Fine-Tuning}

The pretrained models are fine-tuned on the NarrativeQA training dataset using supervised learning. Each input $(s, q)$ is mapped to its corresponding target answer $a$.

Fine-tuning enables the model to adapt to the task-specific distribution, improving its ability to generate concise, relevant, and semantically accurate answers aligned with the dataset.

During this process, the model learns to focus on important parts of the input, such as key events and relevant entities within the story.

This step plays a crucial role in improving baseline performance, as it aligns the pretrained model with the specific requirements of narrative question answering.

\subsection{Self-Ensemble Generation}

Instead of generating a single output, the model produces multiple candidate answers for each input. For a given sample $(s_i, q_i)$, the candidate set is defined as:
\[
A_i = \{a_1, a_2, \dots, a_K\}
\]
where $K = 5$ in our experiments.
The value of K was selected as a trade-off between candidate diversity and computational cost. Increasing K may provide additional answer diversity but also increases inference time. Future work will investigate the sensitivity of the proposed framework to different values of K.

These candidates are generated using stochastic decoding, allowing the model to explore diverse answer formulations.

For candidate generation, stochastic sampling is performed using a temperature value of T = 0.7, which provides a balance between diversity and answer stability.

This diversity increases the likelihood that at least one of the generated answers is correct.

Generating multiple outputs also provides a richer set of candidates that can be evaluated during the selection stage.

This step forms the foundation of the self-ensemble approach, as it enables the model to move beyond single-path inference.

\subsection{Self-Consistency-Based Reranking}

To select the final answer, a semantic similarity-based Self-Consistency-Based mechanism is applied. Pairwise similarity is computed between all candidate answers, and each candidate is assigned a Self-Consistency-Based score:

\[
C(a_m) = \sum_{j \neq m} sim(a_m, a_j)
\]
In our implementation, semantic similarity is computed using
SimCSE sentence embeddings and cosine similarity.

where $sim(a_m, a_j)$ denotes the semantic similarity between two answers.

The final output is selected as:
\[
a^* = \arg\max_{a_m \in A_i} C(a_m)
\]

This approach assumes that the most reliable answer is the one that is most consistent with the other generated candidates.

By relying on semantic similarity rather than exact matching, the method can capture agreement at the meaning level, even when answers differ in wording.

\subsection{Optimization and Evaluation}

Fine-tuning is performed using cross-entropy loss for sequence generation. The model parameters are optimised to maximise the likelihood of the target answer given the input sequence.

This objective encourages the model to generate outputs that are both syntactically correct and semantically aligned with the ground truth.

The framework is evaluated under four settings:

\begin{figure*}[htbp]
\centering
\includegraphics[width=0.9\linewidth,height=4cm]{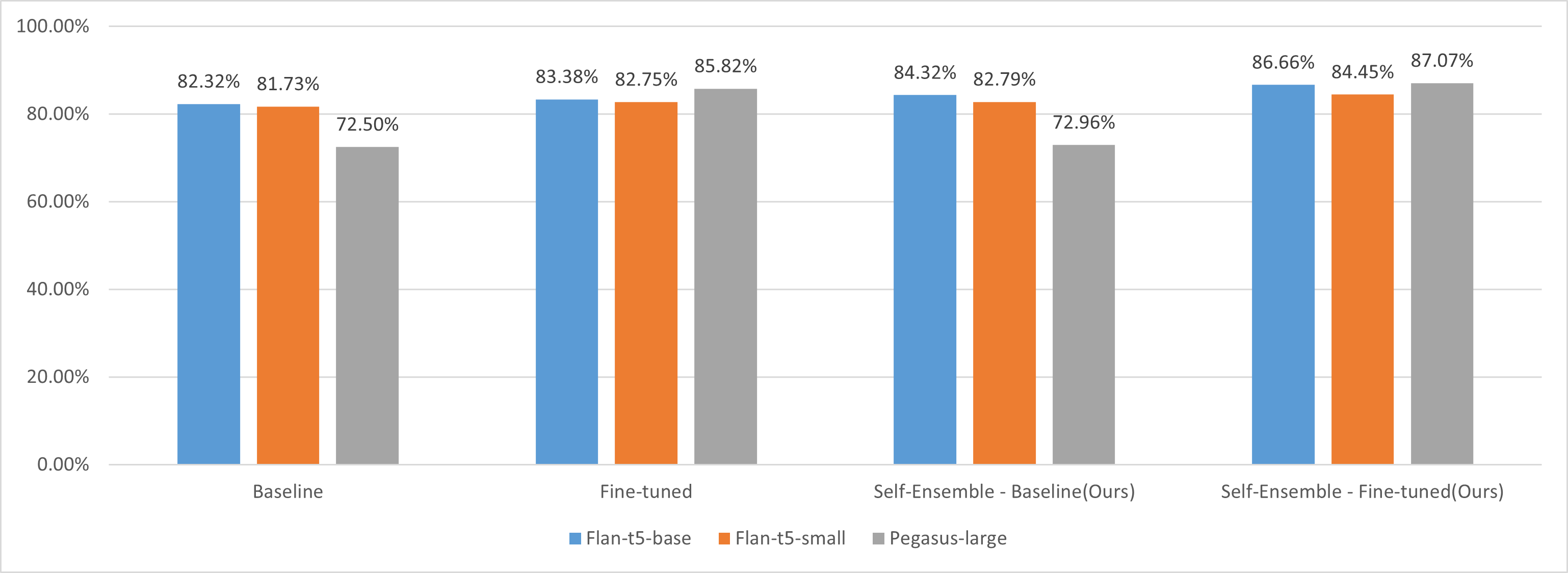}
\caption{Visual comparison of model performance across baseline, fine-tuned, and self-ensemble settings.}
\label{fig:comparison}
\end{figure*}

\begin{itemize}
    \item Baseline: pretrained model with single-answer generation
    \item Fine-tuned: fine-tuned model with single-answer generation
    \item Baseline + Self-Ensemble (Ours)
    \item Fine-tuned + Self-Ensemble (Ours)
\end{itemize}

This evaluation setup allows a clear comparison between standard generation and the proposed self-ensemble strategy.

\subsection{Visualization of Model Performance}

To provide a clearer understanding of the impact of the proposed framework, we present a visual comparison of model performance under different experimental settings, as shown in Fig.~\ref{fig:comparison}.

The figure illustrates the performance of the evaluated models across four configurations: baseline, fine-tuned, self-ensemble baseline, and self-ensemble fine-tuned. Each group represents a specific setting, while the bars correspond to different models.

This visualization highlights the effect of both fine-tuning and the proposed self-ensemble strategy. Fine-tuning improves performance across all models, while the self-ensemble approach further enhances the results by selecting more consistent answers.

Notably, the combination of fine-tuning and self-ensemble yields the highest performance across all models, demonstrating the effectiveness of integrating multi-answer generation with semantic consistency-based selection.

\section{Results and Discussion}

This section presents the experimental evaluation of the proposed Self-Ensemble Self-Consistency-Based reranking framework. The analysis focuses on measuring the impact of multi-answer generation and consensus-based selection compared to standard single-output generation.

\subsection{Benchmarking}

Table~\ref{tab:results} presents the benchmarking results across the evaluated settings. The results compare baseline generation, standard fine-tuning, baseline self-ensemble, and fine-tuned self-ensemble using the selected evaluation metrics.

Evaluation is conducted using BERTScore, which measures semantic similarity between generated answers and reference answers. This metric is particularly suitable for open-ended generation tasks, where exact matching is not sufficient.

Using BERTScore ensures that the evaluation captures semantic correctness rather than surface-level similarity. This provides a more accurate assessment of the model's ability to generate meaningful and relevant answers.

\subsection{Overall Performance}

Table~\ref{tab:results} presents the performance of all evaluated models under four different settings: baseline, fine-tuned, baseline with self-ensemble, and fine-tuned with self-ensemble. The results clearly show that the proposed self-ensemble strategy consistently improves performance across all models.

\begin{table}[htbp]
\caption{Performance comparison under different experimental settings.}
\centering
\small
\setlength{\tabcolsep}{5pt}
\begin{tabular}{lcc|cc}
\toprule
 & \multicolumn{2}{c|}{Single Output} & \multicolumn{2}{c}{Self-Ensemble (Ours)} \\
\cmidrule(r){2-3} \cmidrule(l){4-5}
Model & Baseline & Fine-tuned & Baseline & Fine-tuned \\
\midrule
FLAN-T5-Base   & 82.32\% & 83.38\% & 84.32\% & \textbf{86.66\%} \\
FLAN-T5-Small  & 81.73\% & 82.75\% & 82.79\% & \textbf{84.45\%} \\
Pegasus-Large  & 72.50\% & 85.82\% & 72.96\% & \textbf{87.07\%} \\
\bottomrule
\end{tabular}
\label{tab:results}
\end{table}

Several important observations can be drawn from these results. First, fine-tuning consistently improves performance across all models. This indicates that task-specific adaptation helps align the model with the NarrativeQA data distribution, enabling more accurate and contextually relevant answer generation.

However, the improvement achieved through fine-tuning alone is relatively limited compared to the gains obtained by the proposed self-ensemble strategy. This highlights that model training alone is not sufficient to fully address the variability introduced by stochastic decoding during inference.

A more significant trend is observed when applying self-ensemble reranking. In both baseline and fine-tuned settings, generating multiple candidate answers and selecting the most semantically consistent one leads to consistent performance gains. This demonstrates that answer quality in generative tasks is highly dependent on the selection process, not only on the model's generation capability.

Notably, the improvement in the baseline setting confirms that the proposed method is model-agnostic and does not rely on additional training. This suggests that consensus-based reranking can serve as a lightweight enhancement applicable to pretrained models without modification. In practice, this makes the approach easy to integrate into existing systems without increasing training complexity.

The most substantial improvement is observed in the case of Pegasus-Large, where performance increases from 72.50\% to 87.07\%, representing a gain of +14.57\%. This indicates that models with higher variance in generated outputs benefit more from the proposed approach, as the self-ensemble mechanism effectively filters out inconsistent or low-quality responses. It also suggests that the method is particularly useful for models that produce diverse but unstable outputs.

Furthermore, the combination of fine-tuning and self-ensemble reranking yields the best overall performance across all models. This demonstrates that the two components are complementary. Fine-tuning improves the quality of generated candidates, while self-ensemble reranking enhances the reliability of the final prediction by reducing dependence on a single decoding path.

Another important observation is that the improvements are consistent across different model sizes. This indicates that the proposed approach does not depend on model scale and can be applied effectively to both smaller and larger architectures. Such consistency further supports the generality of the method.

Overall, these findings highlight that improving inference-time decision making is as important as improving model training in generative question answering. The proposed framework provides an effective strategy for enhancing robustness and consistency by leveraging agreement among multiple generated outputs. This also suggests that future improvements in generative models may benefit from focusing not only on better training objectives, but also on more reliable inference and selection mechanisms.

\section{Conclusion }

This paper presented a semantic self-consistency-based reranking framework for narrative question answering. The proposed approach improves inference-time robustness by generating multiple candidate answers and selecting the response with the highest semantic agreement among generated candidates. Experimental results on the NarrativeQA dataset demonstrated consistent improvements across FLAN-T5 and Pegasus models under both pretrained and fine-tuned settings.

The findings suggest that semantic-level agreement can provide a more flexible alternative to traditional exact-match aggregation methods, particularly for open-ended generative tasks where semantically equivalent answers may differ in wording.

A limitation of the current study is that evaluation was conducted primarily using BERTScore. Although BERTScore is suitable for measuring semantic similarity in open-ended generation tasks, additional evaluation metrics and comparison baselines would provide a more comprehensive assessment of the proposed framework.

Despite these promising results, several directions remain for future research. First, future work will include a direct comparison against standard self-consistency baselines such as majority voting and exact-match voting to better quantify the contribution of semantic reranking. Second, an ablation study will be conducted to investigate the impact of the number of generated candidates ($K$), decoding temperature, and alternative semantic similarity metrics. Third, additional evaluation metrics including ROUGE, F1, and Exact Match will be incorporated to enable broader comparison with existing NarrativeQA literature. Finally, future research may explore more advanced verification-based and reasoning-aware reranking strategies, as well as adaptive sampling techniques for improving both efficiency and answer quality.

Overall, the proposed framework demonstrates that improving answer selection during inference can be an effective and lightweight strategy for enhancing generative question answering systems.


\begin{thebibliography}{00}

\bibitem{wang2023selfconsistency}
X. Wang, J. Wei, D. Schuurmans, Q. Le, E. H. Chi, S. Narang, A. Chowdhery, and D. Zhou,
``Self-Consistency Improves Chain-of-Thought Reasoning in Language Models,''
in \textit{Proceedings of the International Conference on Learning Representations (ICLR)}, 2023.

\bibitem{gao2021simcse}
T. Gao, X. Yao, and D. Chen,
``SimCSE: Simple Contrastive Learning of Sentence Embeddings,''
\textit{Proceedings of the 2021 Conference on Empirical Methods in Natural Language Processing (EMNLP)}, 2021.

\bibitem{lightman2023verify}
H. Lightman, V. Kosaraju, Y. Burda, H. Edwards, B. Baker, T. Lee, J. Leike, J. Schulman, I. Sutskever, and K. Cobbe,
``Let’s Verify Step by Step,''
in \textit{Proceedings of the International Conference on Learning Representations (ICLR)}, 2024.



\bibitem{wan2024dynamic}
G. Wan, Y. Wu, J. Chen, and S. Li,
``Dynamic Self-Consistency: Leveraging Reasoning Paths for Efficient LLM Sampling,''
in \textit{International Conference on Learning Representations (ICLR)}, 2024.


\bibitem{zhou2025rpc}
Z. Zhou, Y. Tan, Z. Li, Y. Yao, L.-Z. Guo, X. Ma, and Y.-F. Li,
``Bridging Internal Probability and Self-Consistency for Effective and Efficient LLM Reasoning,''
\textit{arXiv preprint arXiv:2502.00511}, 2025.

\bibitem{wan2025rasc}
G. Wan, Y. Wu, J. Chen, and S. Li,
``Reasoning Aware Self-Consistency: Leveraging Reasoning Paths for Efficient LLM Sampling,''
in \textit{Proceedings of NAACL-HLT}, pp. 3613--3635, 2025.

\bibitem{chen2023usc}
X. Chen, R. Aksitov, U. Alon, J. Ren, K. Xiao, P. Yin, S. Prakash, C. Sutton, X. Wang, and D. Zhou,
``Universal Self-Consistency for Large Language Model Generation,''
\textit{arXiv preprint arXiv:2311.17311}, 2023.

\bibitem{medsyn2024}
G. Kumichev, P. Blinov, Y. Kuzkina, V. Goncharov, G. Zubkova, N. Zenovkin, A. Goncharov, and A. Savchenko,
``MedSyn: LLM-Based Synthetic Medical Text Generation Framework,''
in \textit{ECML PKDD}, 2024.

\bibitem{mo2024detection}
Y. Mo, H. Qin, Y. Dong, Z. Zhu, and Z. Li,
``Large Language Model (LLM) AI Text Generation Detection Based on Transformer Deep Learning Algorithm,''
\textit{arXiv preprint arXiv:2405.06652}, 2024.


\bibitem{molham2026curriculum}
A. Alansary, M. Mohamed, and A. Hamdi,
"A Severity-Based Curriculum Learning Strategy for Arabic Medical Text Generation,"
\textit {arXiv preprint arXiv:2604.06365}, 2026.


\bibitem{molham2026severity}
A. Alansary, M. Mohamed, and A. Hamdi,
"Severity-Aware Weighted Loss for Arabic Medical Text Generation,"
\textit {arXiv preprint arXiv:2604.06346}, 2026.

\bibitem{li2025distributional}
Y. Li, J. Zhang, S. Feng, P. Yuan, X. Wang, J. Shi, Y. Zhang, C. Tan, B. Pan, Y. Hu, and K. Li,
``Revisiting Self-Consistency from Dynamic Distributional Alignment Perspective on Answer Aggregation,''
in \textit{Findings of the Association for Computational Linguistics: ACL 2025}, 2025, pp. 25208--25223.

\bibitem{wang2022selfconsistency}
X. Wang, J. Wei, D. Schuurmans, Q. Le, E. H. Chi, S. Narang, A. Chowdhery, and D. Zhou,
``Self-Consistency Improves Chain-of-Thought Reasoning in Language Models,''
\textit{arXiv preprint arXiv:2203.11171}, 2022.

\bibitem{golestaneh2020consistency}
S. A. Golestaneh and K. M. Kitani,
``Importance of Self-Consistency in Active Learning for Semantic Segmentation,''
\textit{arXiv preprint arXiv:2008.01860}, 2020.

\bibitem{chang2024ensemble}
C.-H. Chang, M. M. Lucas, Y. Lee, C. C. Yang, and G. Lu-Yao,
``Beyond Self-Consistency: Ensemble Reasoning Boosts Consistency and Accuracy of LLMs,''
\textit{arXiv preprint arXiv:2404.13149}, 2024.

\bibitem{xu2025selfensemble}
Z. Xu, G. Wang, G. Zheng, Y.-N. Chuang, A. Szalay, X. Hu, and V. Braverman,
``Self-Ensemble: Mitigating Confidence Distortion for Large Language Models,''
\textit{Findings of the Association for Computational Linguistics: EMNLP}, pp. 16603--16615, 2025.


\bibitem{wang2025srnle}
Y. Wang and P. Atanasova,
``Self-Critique and Refinement for Faithful Natural Language Explanations,''
in \textit{Proceedings of the 2025 Conference on Empirical Methods in Natural Language Processing (EMNLP)}, 2025.


\bibitem{hawkins2021consistency}
C. Hawkins, V. N. Ioannidis, S. Adeshina, and G. Karypis,
``Scalable Consistency Training for Graph Neural Networks via Self-Ensemble Self-Distillation,''
in \textit{Proceedings of the Web Conference (WWW)}, 2021.

\bibitem{samanta2026maca}
A. Samanta, A. Magesh, R. Wu, A. Jain, Y. Yu, D. Jiang, B. Vidolov, P. Sajda, Y. Efroni, and K. Hassani,
``Self-Improvement of Language Models by Post-Training on Multi-Agent Debate,''
2026.

\bibitem{ahmed2023selfconsistency}
T. Ahmed and P. Devanbu,
``Better Patching Using LLM Prompting, via Self-Consistency,''
in \textit{Proceedings of the IEEE/ACM International Conference on Software Engineering Workshops (ICSE Workshops)}, 2023.


\bibitem{aloraini2025lexisem}
E. Aloraini, H. Kassab, A. Hamdi, and K. Shaban,
``LexiSem: A Re-ranker Balancing Lexical and Semantic Quality for Enhanced Abstractive Summarization,''
\textit{Neurocomputing}, vol. 650, p. 130816, 2025.

\bibitem{elewa2025balancing}
M. Elewa, A. Hamdi, H. Kassab, and K. Shaban,
``Balancing Factual Consistency and Diversity in Abstractive Summarization via Model-Agnostic Composite Reranking,''
in \textit{2025 IEEE/ACS 22nd International Conference on Computer Systems and Applications (AICCSA)}, 2025, pp. 1--8.

\bibitem{hamdi2024riro}
A. Hamdi, H. Kassab, M. Bahaa, and M. Mohamed,
``Riro: Reshaping Inputs, Refining Outputs Unlocking the Potential of Large Language Models in Data-Scarce Contexts,''
in \textit{The International Conference of Advanced Computing and Informatics}, 2024, pp. 69--79.

\bibitem{elgabry2025confidence}
M. Elgabry and A. Hamdi,
``Confidence-Credibility Aware Weighted Ensembles of Small LLMs Outperform Large LLMs in Emotion Detection,''
in \textit{International Conference of Reliable Information and Communication Technology}, 2025, pp. 170--179.

\bibitem{hamdi2025ensemble}
A. Hamdi, M. Mohamed, R. Emad, and K. Shaban,
``An Ensemble Classification Approach in A Multi-Layered Large Language Model Framework for Disease Prediction,''
in \textit{2025 IEEE/ACS 22nd International Conference on Computer Systems and Applications (AICCSA)}, 2025, pp. 1--5.

\bibitem{basem2025two}
M. Basem, I. Oshallah, A. Hamdi, K. Shaban, and H. Kassab,
``Two-Stage Quranic QA via Ensemble Retrieval and Instruction-Tuned Answer Extraction,''
in \textit{2025 IEEE/ACS 22nd International Conference on Computer Systems and Applications (AICCSA)}, 2025, pp. 1--8.

\bibitem{akl2025fusion}
A. Akl and A. Hamdi,
``Fusion Strategies for Embedding Models: Enhancing Text Representations Across MTEB Tasks Through Lightweight and Trainable Ensembles,''
in \textit{2025 Intelligent Methods, Systems, and Applications (IMSA)}, 2025, pp. 126--131.

\bibitem{hamdi2024llm}
A. Hamdi, A. A. Mazrou, and M. Shaltout,
``Llm-Sem: A Sentiment-Based Student Engagement Metric Using LLMs for E-Learning Platforms,''
in \textit{The International Conference of Advanced Computing and Informatics}, 2024, pp. 145--154.

\bibitem{wassim2024llm}
L. Wassim, K. Mohamed, and A. Hamdi,
``Llm-Daas: LLM-Driven Drone-as-a-Service Operations from Text User Requests,''
in \textit{The International Conference of Advanced Computing and Informatics}, 2024, pp. 108--121.

\bibitem{abdellaif2024erpa}
O. H. Abdellaif, A. N. Hassan, and A. Hamdi,
``Erpa: Efficient RPA Model Integrating OCR and LLMs for Intelligent Document Processing,''
in \textit{2024 International Mobile, Intelligent, and Ubiquitous Computing Conference (MIUCC)}, 2024, pp. 295--300.
\end{thebibliography}
\end{document}